\pdfoutput=1

\documentclass[11pt]{article}

\usepackage[final]{ACL2023}

\usepackage{times}
\usepackage{latexsym}
\usepackage{graphicx}

\usepackage[T1]{fontenc}

\usepackage[utf8]{inputenc}

\usepackage{microtype}
\usepackage{amsmath} 

\usepackage{inconsolata}

\title{Towards Better Serialization of Tabular Data for Few-shot Classification with Large Language Models}

\author{%
  Sukriti Jaitly \quad Tanay Shah \quad Ashish Shugani   \quad Razik Singh Grewal  \\
  Carnegie Mellon University \\
  \texttt{\{sjaitly,tashah,ashugani,rgrewal\}@andrew.cmu.edu}
}

\begin{document}
\maketitle
\begin{abstract}
We present a study on the integration of Large Language Models (LLMs) in tabular data classification, emphasizing an efficient framework. Building upon existing work done in TabLLM \cite{hegselmann2023tabllm}, we introduce three novel serialization techniques, including the standout LaTeX serialization method. This method significantly boosts the performance of LLMs in processing domain-specific datasets, Our method stands out for its memory efficiency and ability to fully utilize complex data structures. Through extensive experimentation, including various serialization approaches like feature combination and importance, we demonstrate our work's superiority in accuracy and efficiency over traditional models.
\end{abstract}

\section{Introduction}
Advancements in deep learning have significantly impacted fields like natural language processing and computer vision, but tabular data hasn't seen a parallel rise \cite{borisov2022deep}. This disparity stems from tabular data's unique characteristics, such as mixed data types and fewer columns.

Large language models (LLMs) like GPT-3 \cite{brown2020language}, with their extensive pre-training, have shown prowess in tasks ranging from few-shot text classification to tabular data cleaning. Recognizing this potential, TabLLM was introduced as a bridge between LLMs and tabular data classification, leveraging serialized natural-language representations of tabular rows for classification. Initial evaluations highlighted TabLLM's competitive edge over other models, including traditional ones like gradient-boosted trees \cite{hegselmann2023tabllm}.

Recognizing the potential of LLMs in tabular data classification, the TabLLM framework was introduced. This framework leverages serialized natural-language representations of tabular rows for classification, and has shown a competitive edge over traditional models like gradient-boosted trees \cite{hegselmann2023tabllm}. Building upon the success of TabLLM, our study introduces a novel framework, which enhances the performance of LLMs in tabular data classification through advanced serialization methods. Our approach significantly diverges from previous models by utilizing LaTeX serialization, which improves memory efficiency and the model's ability to process complex data structures.

Our project delves into various serialization approaches, including feature combination and importance, to demonstrate the superiority of our techniques in terms of accuracy and efficiency. We also explore the integration of key tabular features into the LLM framework, enhancing its interpretability and effectiveness in classification tasks. This study presents a comprehensive examination of using novel serialisation techniques in tabular data classification, highlighting its potential to transform data processing across various industry areas.

\section{Related Work}

\textbf{TabLLM:} 
TabLLM is a novel framework for few-shot classification of tabular data using large language models (LLMs). This approach focuses on converting tabular data into natural language representations, then employing LLMs, particularly the T0 model and GPT-3, for classification tasks. This method addresses challenges in obtaining sufficient labeled data for traditional supervised learning, especially in fields like healthcare.

It explores different methods for serializing tabular data into natural language inputs for language models, focusing on improving zero and few-shot classification performance. Nine serialization formats are studied, ranging from simple list templates to more complex methods like Table-To-Text and GPT-3 based serializations. The paper also examines ablations like listing only values or permuting names/values to understand the impact of column names and the precision of feature values on classification performance.

TabLLM outperforms traditional deep-learning methods and rivals gradient-boosted trees, particularly in few-shot learning scenarios. TabLLM's efficacy is highlighted by its ability to leverage the extensive knowledge encoded in pre-trained LLMs, requiring minimal labeled data. The framework's adaptability to different types of tabular data and its competitive performance against traditional models underline its potential in domains where labeled data is scarce or expensive to obtain.

TabLLM requires significant computational resources, including large GPUs for fine-tuning, and its performance may decline if a dense feature set exceeds an LLM's token limit. Despite these drawbacks, TabLLM shows excellent performance in tabular classification, especially in few-shot scenarios, outperforming algorithms like XGBoost and SAINT.
\\
\\
\textbf{TabNet:} TabNet is a novel high-performance and interpretable canonical deep tabular data learning architecture which uses sequential attention to choose which features to reason
from at each decision step, enabling interpretability and more
efficient learning as the learning capacity is used for the most
salient features. It stands out from its competitors in the following ways:\\
1. It is built to work on raw tabular data without the necessity for preprocessing, allowing for gradient descent-based optimization and seamless integration with end-to-end learning.\\
2. A distinctive feature of TabNet is its sequential attention mechanism, which determines the features to focus on during each decision step.\\
3. TabNet's design choices endow it with commendable advantages. Empirical studies reveal that it either surpasses or matches the performance of other tabular learning models across diverse datasets and problem types. \\
4. Another pioneering feature of TabNet is its capability to harness the power of unsupervised pre-training for tabular data, which leads to remarkable performance enhancements by predicting masked features.

TabLLM and TabNet are both related works in the field of tabular data analysis and machine learning, but they approach the problem from different angles and have distinct characteristics. While both TabLLM and TabNet deal with tabular data, they differ in their core objectives and methodologies. TabLLM emphasizes few-shot classification using large language models, making it suitable for scenarios with limited labeled data. In contrast, TabNet prioritizes interpretability and efficient learning, focusing on creating a versatile and competitive deep learning model for tabular data. Their relatedness lies in their common goal of advancing the capabilities of machine learning in the domain of tabular data analysis, but they take distinct paths to achieve it.

\section{Method}

\subsection{Serialization}

Leveraging Large Language Models (LLMs) for tabular data necessitates converting tables into a format akin to natural language. This procedure involves serializing table content into coherent text while incorporating task-specific cues into the prompt. Our approach tested various serialization techniques, emphasizing the significance of features, their combinations, and structuring the input with a LaTeX framework to align with LLMs' textual processing capabilities.

\subsection{Fine-tuning}

We implemented fine-tuning on the T0 (3 billion) model \cite{sanh2021multitask} using the Intrinsic Attention-based Prompt Tuning (IA3) \cite{liu2022fewshot} as a Parameter-Efficient Fine-Tuning (PEFT) technique. The approach was applied to the serialized dataset across a broad spectrum of training examples, ranging from zero-shot to 32 shots. This detailed exploration covered various levels of data scarcity, allowing us to rigorously evaluate the model's adaptability and performance across a gradient of data availability scenarios.

\subsection{LLM for Prediction}

 Given an LLM with a vocabulary $V$, the output prompted by the LLM, denoted as $\text{LLM}(\text{serialize}(F, x), p) \in V^*$, is derived from the natural-language input serialization of the features $F$ and instance $x$ combined with a prompt $p$. In a few-shot learning context, the set $\{(\text{serialize}(F, x), p) \mid (x, y) \in D_k \}$ is utilized for fine-tuning the LLM. The generated text in the vocabulary space $V^*$ must then be translated into a valid class label in $C$. Existing methods, such as the verbalizer approach, establish this mapping from LLM output tokens to discrete label spaces. We adopt a manual specification for this mapping in our work.

\begin{figure*}[ht]
    \centering
    \includegraphics[width=16cm]{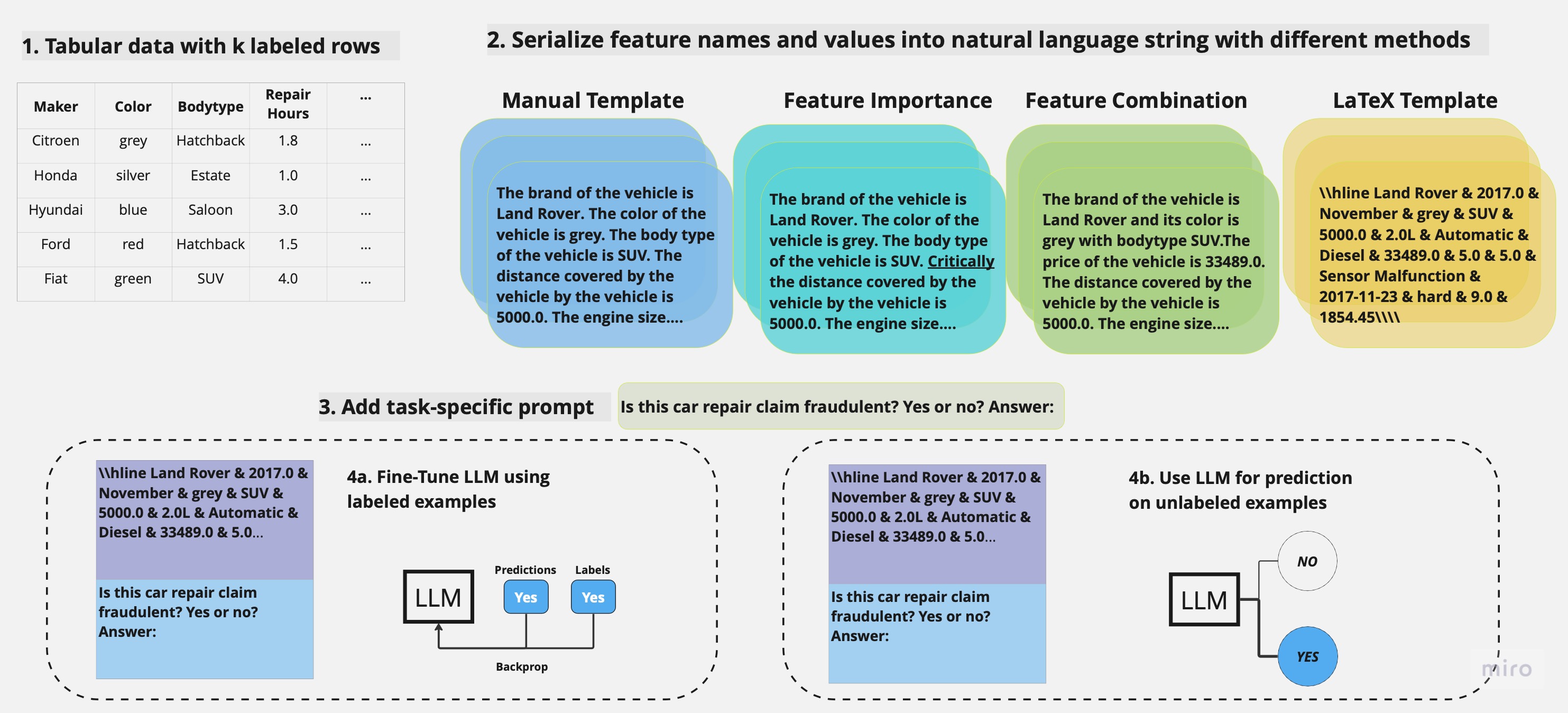}
    \caption{Overview of Framework}
    \label{fig:yourlabel}
\end{figure*} 

\section{Experimental Setup}
\subsection{Dataset Details}
In our research, we utilized the Vehicle Claims dataset, introduced by \cite{chawda2022unsupervised} in "Unsupervised Anomaly Detection for Auditing Data and Impact of Categorical Encodings". This dataset  focuses on fraudulent insurance claims in the automotive sector, characterized by a high number of categorical attributes. It includes detailed information on vehicle make, model, color, registration year, body type, engine size, gearbox, fuel type, and various numerical attributes like mileage, price, seat and door numbers, along with the type and specifics of vehicle issues, repair complexity, duration, and cost. The dataset is available on GitHub.

\subsection{Baseline}
TabLLM investigated multiple approaches to serialize tabular data for language model processing, with the Text-template method emerging as the most effective. Therefore, TabLLM with Text-template serialization was set as a baseline. The Text Template serialization in TabLLM involves converting features and values into a natural language string where each column name is followed by its corresponding value. This format mimics how humans might naturally describe the data.

For example, a serialized entry could read, "Age is 35, sex is female, race is Caucasian," and so on, adding relevant visit information like date, doctor type, and primary complaint. This method respects the token limit of the LLM, ensuring that a manageable amount of information is processed.

In the test AUC performance results, TabLLM generally outperforms or is competitive with XGBoost and TabPFN across various datasets (serialized using Text-template) and numbers of shots, especially in the very-few-shot regime. 
\subsection{Serialization with Feature Combination}

In refining TabLLM, we introduced a feature combination technique, moving away from the traditional single-feature-per-sentence format. This new approach is designed to align closely with natural language, enhancing the model's ability to interpret and comprehend the complex interrelationships between different features in tabular data. By transitioning from a simplistic serialization template, which treated each feature in an isolated way, to a more integrated format, we mimic the natural flow of human language. 
\\
For example, in our vehicle claims dataset, attributes like make, color and body type are now combined into a single richer sentence. 

\subsection{Serialization with Feature Importance}

A critical part of our approach involves understanding the significance of different features in our dataset. To achieve this, we employed a covariance-based method to identify the most influential features with respect to the target variable, 'Label'. The process involved the following steps:

\begin{enumerate}
    \item Conversion of categorical data into numerical format using one-hot encoding.
    \item Computation of covariance values between all features and the target variable.
    \item Ranking of features based on the absolute values of their covariance and selection of the top 4.
    \item Two techniques were then explored for enhancing feature importance recognition:
    \begin{enumerate}
        \item Adding the prefix “Critically” to the most important features of the prompt.
        \item Adding a sentence at the end of the original prompt as: “The 4 most important features for the inference are: A, B, C, D.”
    \end{enumerate}
    
\end{enumerate}

This method provided us with a clearer picture of which features our model should prioritize during classification, in an attempt to enhance the accuracy and efficiency of the LLM in handling tabular data.

\subsection{LaTeX Serialization}
Moving away from the traditional table-to-text serialization techniques, we explored the novel approach of converting tabular data into LaTeX code format. This LaTeX representation of the table was then used as the input for fine-tuning our LLM by just passing a row representation preceded by hline tag without any headers. This method is grounded in the hypothesis that such a structured representation, familiar in academic and scientific publications, might aid the LLM in capturing the nuances of tabular data more effectively. The LaTeX format, being more structured and consistent than plain text, could potentially allow the LLM to understand and process the tabular data with higher accuracy, especially in complex dataset where traditional serialization methods may fall short.

\section{Results}

\begin{table*}[h]
\renewcommand{\arraystretch}{1.5}
\centering
\begin{tabular}{|c|ccccccccc|}
\hline
\textbf{TabLLM} & \multicolumn{9}{|c|}{\textbf{Number of Shots}} \\ \hline
\textbf{Serialization Method} & \textbf{0} & \textbf{4} & \textbf{8} & \textbf{16} & \textbf{32} & \textbf{64} & \textbf{128} & \textbf{256} & \textbf{512} \\
\hline
XGBoost & \_ & 0.5 & 0.5 & 0.441 & 0.617 & 0.579 & 0.800 & 0.912 & 0.946 \\
\hline
Text Template (\textbf{Baseline}) & 0.485 & 0.495 & 0.512 & 0.557 & 0.642 & 0.954 & 0.985 & \textbf{0.994} & \textbf{0.997} \\
\hline
Feature Importance & 0.477 & 0.491 & 0.499 & 0.54 & 0.653 & 0.942 & 0.944 & 0.972 & 0.982 \\
\hline
Feature Combination & 0.509 & 0.521 & 0.531 & 0.582 & 0.665 & 0.933 & \textbf{0.986} & 0.992 & 0.997 \\
\hline
LaTeX & \textbf{0.686} & \textbf{0.684} & \textbf{0.652} & \textbf{0.666} & \textbf{0.778} & \textbf{0.962} & \textbf{0.986} & 0.990 & 0.992\\
\hline
\end{tabular}
\caption{AUC for varying number of training examples (shots)}
\label{nshotVSauc}
\end{table*}

\begin{figure}[h]
\includegraphics[width=7.75cm, height=6cm]{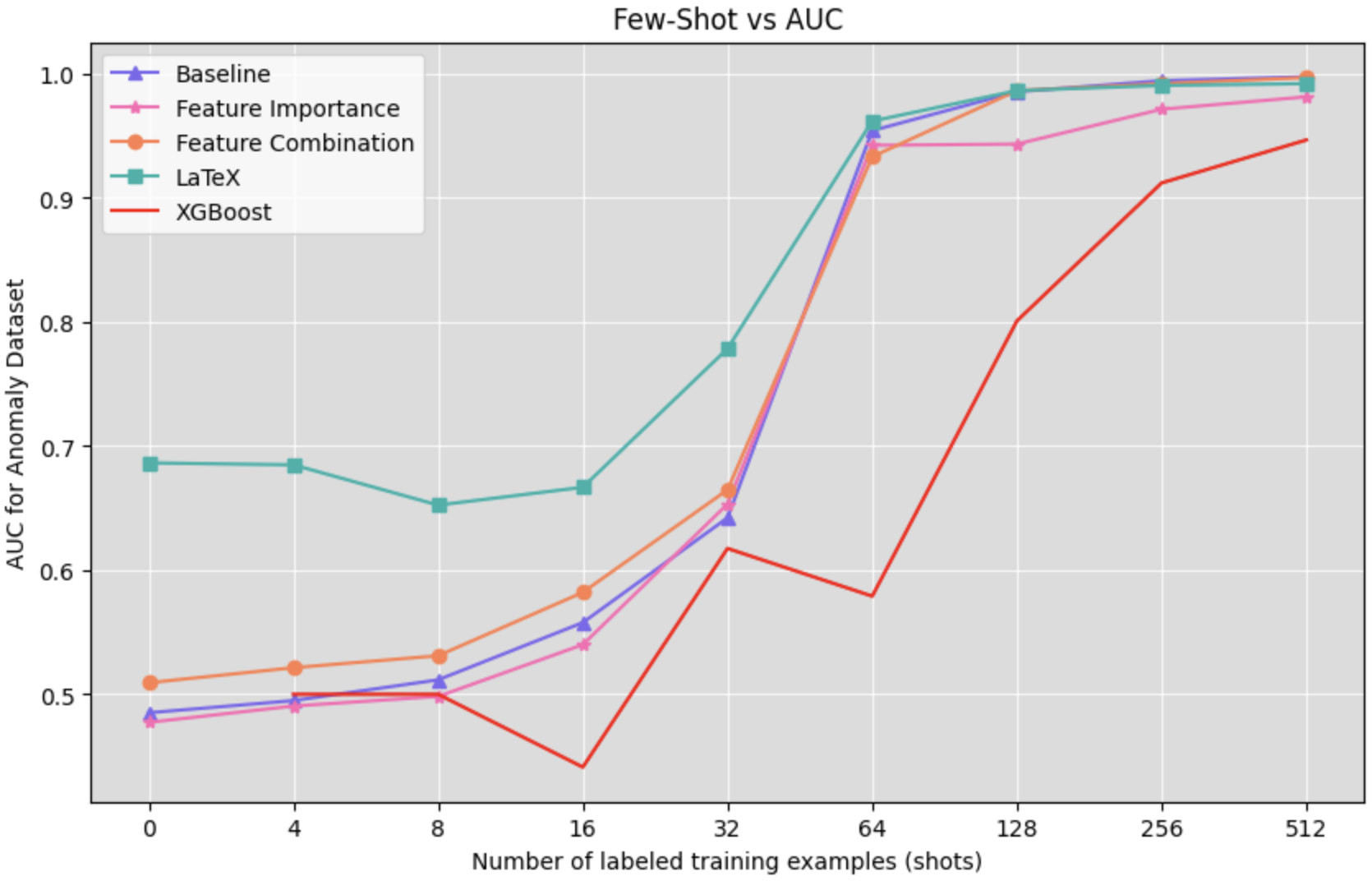}
\caption{AUC for the anomaly dataset with different number of training examples (shots)}
\label{fignshotVSauc}
\end{figure}

Figure \ref{fignshotVSauc} shows the performance of various serialization techniques compared with the traditional Machine Learning model, XGBoost \cite{Chen_2016}, and the best performing serialization technique presented in TabLLM. In zero-shot setting LaTeX serialization performs significantly well when compared to other three serialization techniques that performs similar to the XGBoost.

Although, the LaTeX serialization shows small dip in performance with additional in-context examples, it still outperforms other methods. Our hypothesis for the dip in performance is that the selection of examples is not the best set of examples for the model to learn. As we increase the number of shots, we see an increase in the performance with all the techniques and after 64-shots, we get similar performance. These results are also show in Table \ref{nshotVSauc}.

Notably, we were constrained to a maximum of 24 GB of GPU when utilizing the highest available AWS instance (g5.2xlarge). Due to this constraint and the memory usage of T03B model, when using all 17 columns with serialization with manual template, feature importance and feature combination we observer that after 32-shots we face CUDA Out of Memory error. To mitigate this problem, we eliminated few trivial columns based on feature importance from the dataset and experimented with the same. However, while working with LaTeX serialization, we used a single row represented by hline tag and the template did not contain any headers related to the table due to which we did not face the memory issue and hence we were able to fit all the columns. Hence, LaTeX serialization not only gives better performance but also provides memory efficiency.

\section{Discussion}
One of the most interesting aspects of our project is our approach to model tuning. We employed the parameter-efficient fine-tuning technique. Initially, we hadn't anticipated the need for such a technique, but it became evident that to draw parallels with TabLLM behavior and draw results with the given computational budget, we have to employ those techniques. 
\\
A significant challenge we encountered was the computational restrictions, which posed limitations on the scale of our dataset. Tuning the model with our chosen dataset proved to be more demanding than we had anticipated. As a result, the experiments we have conducted so far are based on a relatively small subset of data. This limitation is crucial to note, as the results derived from such a limited dataset might not be comprehensive and that means that further research and evaluations on more extensive datasets are necessary to validate and generalize our results.
\section{Conclusion}
This study has successfully demonstrated the innovative application of Large Language Models (LLMs) in the domain of tabular data classification, with a specific focus on our new LaTeX serialisation framework. Our approach, which builds upon the foundational TabLLM \cite{hegselmann2023tabllm} techniques, has introduced novel serialization methods that have proved effective in handling domain-specific datasets. Among these, the LaTeX serialization method stands out, leveraging a structured LaTeX code format for tabular data representation. This method not only enhances LLM performance in classification tasks but also exhibits remarkable memory efficiency and computational efficacy, enabling the complete utilization of all data columns without encountering computational constraints like CUDA Out of Memory errors. Additionally, our exploration of other serialization techniques, such as integrating multiple features into coherent sentences and emphasizing feature importance, has furthered our understanding of LLMs' capabilities in interpreting complex data relationships. While these methods played a secondary role to the primary focus on LaTeX serialization, they collectively contribute to a more nuanced understanding of how LLMs can be effectively applied to diverse data scenarios. This study, therefore, not only underscores the potential of the LaTeX serialization in the realm of LLM-driven tabular data analysis but also sets the stage for future research in developing more sophisticated and efficient tools for data processing across various industry domains.

\section{Societal Implications}
The application of LLMs in tabular data classification has far-reaching implications in various societal domains. By enhancing the accuracy and efficiency of such models, we can significantly improve decision-making processes in areas like healthcare, finance, and environmental studies. The ability of LLMs to adapt to different domains and handle sparse data can lead to more informed and data-driven solutions to critical societal issues.

\bibliography{custom}
\bibliographystyle{acl_natbib}

\appendix

\end{document}